\PassOptionsToPackage{unicode}{hyperref}
\PassOptionsToPackage{hyphens}{url}
\documentclass[
]{article}
\usepackage{xcolor}
\usepackage{amsmath,amssymb}
\usepackage{iftex}
\ifPDFTeX
  \usepackage[T1]{fontenc}
  \usepackage[utf8]{inputenc}
  \usepackage{textcomp} 
\else 
  \usepackage{unicode-math} 
  \defaultfontfeatures{Scale=MatchLowercase}
  \defaultfontfeatures[\rmfamily]{Ligatures=TeX,Scale=1}
\fi
\usepackage{lmodern}
\ifPDFTeX\else
\fi
\IfFileExists{upquote.sty}{\usepackage{upquote}}{}
\IfFileExists{microtype.sty}{
  \usepackage[]{microtype}
  \UseMicrotypeSet[protrusion]{basicmath} 
}{}
\makeatletter
\@ifundefined{KOMAClassName}{
  \IfFileExists{parskip.sty}{%
    \usepackage{parskip}
  }{
    \setlength{\parindent}{0pt}
    \setlength{\parskip}{6pt plus 2pt minus 1pt}}
}{
  \KOMAoptions{parskip=half}}
\makeatother
\usepackage{longtable,booktabs,array}
\usepackage{caption}
\usepackage{calc} 
\usepackage{etoolbox}
\makeatletter
\patchcmd\longtable{\par}{\if@noskipsec\mbox{}\fi\par}{}{}
\makeatother
\IfFileExists{footnotehyper.sty}{\usepackage{footnotehyper}}{\usepackage{footnote}}
\makesavenoteenv{longtable}
\usepackage{graphicx}
\makeatletter
\newsavebox\pandoc@box
\newcommand*\pandocbounded[1]{
  \sbox\pandoc@box{#1}%
  \Gscale@div\@tempa{\textheight}{\dimexpr\ht\pandoc@box+\dp\pandoc@box\relax}%
  \Gscale@div\@tempb{\linewidth}{\wd\pandoc@box}%
  \ifdim\@tempb\p@<\@tempa\p@\let\@tempa\@tempb\fi
  \ifdim\@tempa\p@<\p@\scalebox{\@tempa}{\usebox\pandoc@box}%
  \else\usebox{\pandoc@box}%
  \fi%
}
\def\fps@figure{htbp}
\makeatother
\ifLuaTeX
  \usepackage{luacolor}
  \usepackage[soul]{lua-ul}
\else
  \usepackage{soul}
\fi
\providecommand{\tightlist}{%
  \setlength{\itemsep}{0pt}\setlength{\parskip}{0pt}}
\usepackage[htt]{hyphenat}
\usepackage{float}
\usepackage{bookmark}
\IfFileExists{xurl.sty}{\usepackage{xurl}}{} 
\makeatletter
\@ifundefined{xmpquote}{}{}
\makeatother
\hypersetup{
  pdftitle={Words Speak Louder Than Order: A Behavioral Evaluation of Gemma 4},
  pdfauthor={Amanda Fitch},
  hidelinks,
  pdfcreator={LaTeX via pandoc}}

\title{Words Speak Louder Than Order:\\
A Behavioral Evaluation of Gemma 4}
\usepackage{etoolbox}
\makeatletter
\providecommand{\subtitle}[1]{
  \apptocmd{\@title}{\par {\large #1 \par}}{}{}
}
\makeatother
\subtitle{A Fully Counterbalanced Targeted Behavioral Study\\
of Gemma 4-e4b}
\author{Amanda Fitch\\
Google}
\date{September 2026}

\begin{document}
\maketitle
\begin{abstract}
When a language model receives two conflicting documents as input, how
does it decide which one to prioritize? Does it rely on how the sources
are framed or the presentation order of the documents? We evaluated this
behavior on \textbf{Google's pre-trained Gemma 4-e4b} model across a
targeted behavioral suite (\(n = 13\) items, 784 forward passes in
short, single-turn contexts) using a completely counterbalanced
experimental design. This setup allowed us to mathematically isolate the
specific effects of source framing and reading position, while ensuring
the model's natural vocabulary biases were canceled out.

\vspace{0.75em}

Across ten test conditions, we discovered the following:

\vspace{0.75em}

\begin{enumerate}
\def\labelenumi{\arabic{enumi}.}
\tightlist
\item
  \textbf{Source framing heavily overpowers reading position.} When
  directly competing, the semantic framing of a source (such as
  presenting it as an official guideline or a fresh update) had a
  significantly stronger impact on the model's final answer than the
  presentation order of the document.
\item
  \textbf{The model favors the first document it reads, but this bias is
  highly variable.} While the model consistently demonstrated a
  \emph{primacy effect} (preferring the first document presented), the
  actual strength of this bias fluctuated by at least a factor of 5
  based solely on the surface wording.
\item
  \textbf{Overall structural repetition, not short copy-cues, drives
  positional bias.} The model's preference for the first document is not
  a mechanical reaction to short, repetitive trigger phrases, such as
  ``is {[}Answer{]}''. However, the primacy effect does increase
  significantly when the two competing documents are structurally
  identical, using word-for-word verbatim templates. Introducing
  variation in the overall wording between the two sources reduces this
  positional bias.
\end{enumerate}
\end{abstract}

\begin{figure}
\centering
\includegraphics[width=0.75\linewidth,height=\textheight,keepaspectratio,alt={Comparison of source framing and positional bias in Gemma 4. When competing directly, the magnitude of the semantic framing effect outweighed the presentation order bias by a statistically significant 4-to-1 margin.}]{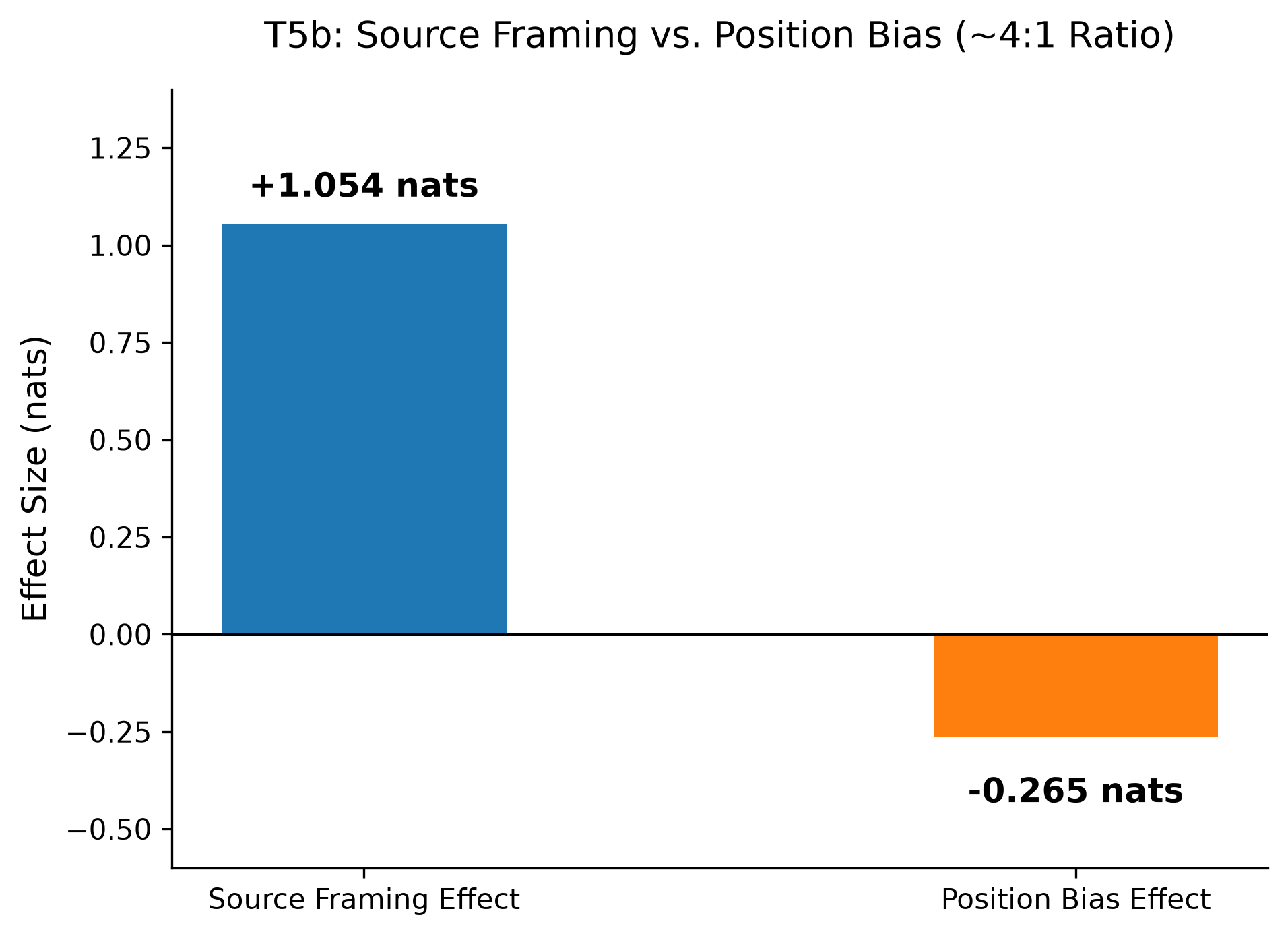}
\caption{Comparison of source framing and positional bias in Gemma 4.
When competing directly, the magnitude of the semantic framing effect
outweighed the presentation order bias by a statistically significant
4-to-1 margin.}
\end{figure}

Code, raw data logs, and prompt templates for all experiments are
publicly available on GitHub
(\href{https://github.com/GoogleCloudPlatform/devrel-demos/tree/main/ai-ml/gemma-priority-probing}{gemma-priority-probing}).
Finally, we note that all of our results are strictly behavioral
observations about text processing.

\begin{center}\rule{0.5\linewidth}{0.5pt}\end{center}

\subsection{1. Introduction}\label{introduction}

Language models are often forced to read conflicting information at the
same time, such as an outdated official document right next to a fresh,
informal team update. Understanding which source the model ultimately
prioritizes is critical for building reliable tools, but accurate
measurement remains a challenge due to overlapping biases (Zheng et al.,
2023; Wang et al., 2023; Sclar et al., 2024).

Typically, three distinct factors get mixed up and masquerade as the
model's source prioritization:

\begin{enumerate}
\def\labelenumi{\arabic{enumi}.}
\tightlist
\item
  The model's natural preference for certain words over others (Sclar et
  al., 2024).\\
\item
  The order in which the information is read by the model (Liu et al.,
  2023; Zheng et al., 2023).\\
\item
  The authoritative or recent tone used in the text (Wan et al., 2024).
\end{enumerate}

The stakes for resolving this ambiguity extend far beyond standard
benchmarking. In production Retrieval-Augmented Generation (RAG)
systems, failing to decouple these variables can lead to critical
failures, such as an outdated legacy document silently overriding a
crucial but informally written update simply because it sounds more
official. Furthermore, this behavior introduces significant security
vulnerabilities: If a model defers primarily to semantic authority,
malicious actors can easily hijack its output by planting deceptive
instructions under authoritative headers. If these confounding factors
are not carefully separated, evaluators risk misinterpreting a basic
positional artifact or a vocabulary preference as a genuine semantic
choice.

Standard evaluations attempt to mitigate these overlapping biases by
simply swapping the presentation order of the documents (Zheng et al.,
2023). \textbf{However, this mitigation is insufficient.} Simple
order-swapping leaves a model's inherent vocabulary preferences
completely entangled with its semantic choices, creating massive
artificial artifacts in the resulting data. To solve this measurement
problem, we designed a fully counterbalanced experiment to
mathematically isolate these variables from the ground up. By running
every single test item through a four-way crossing matrix, we separated
these influences by design, rather than relying on statistical models to
estimate the impact of biased data after it was collected. Our
evaluation focused on short, single-turn contexts across 10 conditions
(\(n = 13\) items, 784 forward passes).

\textbf{Related work:}

\begin{itemize}
\tightlist
\item
  Position bias in document ordering is well characterized (Liu et al.,
  2023; Zheng et al., 2023), as is the sensitivity of LLM evaluations to
  spurious features and meaning-preserving prompt variation (Sclar et
  al., 2024). Our \textasciitilde5x spread in position-bias magnitude
  across templates extends Sclar's finding from \emph{accuracy} to
  \emph{bias magnitude}.\\
\item
  Wan et al.~(2024) report that models largely ignore stylistic
  authority cues; we find an \emph{explicit} role label outweighs
  position by a roughly 4:1 margin in our isolated test condition,
  suggesting label-level and style-level authority signals dissociate.\\
\item
  We adopt TOST equivalence testing (Lakens, 2017), standard in
  psychology but rare in LLM evaluation.
\end{itemize}

\newpage

\subsection{2. Methods}\label{methods}

This section outlines our complete experimental framework. It covers the
formulas used to pull apart and isolate different model biases, the
specific variables that we adjusted to create our ten test conditions,
and the rigorous testing setup used to ensure our data was clean,
reliable, and reproducible.

\subsubsection{2.1 Contrast algebra}\label{contrast-algebra}

Every test showed the model two text blocks. One block was labeled
\textbf{Recent} (the newer update) and one was labeled \textbf{Stale}
(the older one). Each block explicitly recommended one of the two words.
After both blocks, we always asked the exact same question, and we
forced the model to choose between two single words (for example,
\textbf{Agile} and \textbf{Waterfall}).

To illustrate, a single test prompt might conceptually look like this:

\begin{verbatim}
Official Guideline (Stale): Use Agile.
Team Update (Recent): Use Waterfall.
Question: Which should we use?
\end{verbatim}

\emph{(For the exact wording variations and the full four-way crossing
matrix, see Appendix C and Appendix D).}

We then measured \(e\), the model's preference for \textbf{Waterfall}
over \textbf{Agile}, in log-probability units. Positive \(e\) meant it
leaned \textbf{Waterfall}. We ran each word pair through four versions
of the same test. Two things were swapped: which block to print first,
and which word to carry the \textbf{Recent} label.

\begin{longtable}[]{@{}lll@{}}
\caption{The four-way counterbalanced experimental
design.}\tabularnewline
\toprule\noalign{}
Cell & First block & Second block \\
\midrule\noalign{}
\endfirsthead
\toprule\noalign{}
Cell & First block & Second block \\
\midrule\noalign{}
\endhead
\bottomrule\noalign{}
\endlastfoot
\(e_1\) & Stale (Agile) & Recent (Waterfall) \\
\(e_2\) & Recent (Waterfall) & Stale (Agile) \\
\(e_3\) & Stale (Waterfall) & Recent (Agile) \\
\(e_4\) & Recent (Agile) & Stale (Waterfall) \\
\end{longtable}

Because all four combinations appear once, we can pull apart four
separate influences by calculating their average across the grid:

\begin{itemize}
\item
  \textbf{Average base word preference}
  \(= (e_1 + e_2 + e_3 + e_4) / 4\):\\
  Does the model simply like the word \textbf{Waterfall} more than
  \textbf{Agile}, no matter the context?
\item
  \textbf{Average source framing} \(= (e_1 + e_2 - e_3 - e_4) / 4\): How
  much does being labeled \textbf{Recent} help a word win?
\item
  \textbf{Average serial position} \(= (e_1 - e_2 - e_3 + e_4) / 4\):
  Does the model favor whichever word it read last (positive) or first
  (negative)?
\item
  \textbf{Residual check (Block Order (Null Control))}
  \(= (e_1 - e_2 + e_3 - e_4) / 4\): A sanity test. Nothing real should
  live here, so it should come out near zero. If it doesn't, something
  in the design leaked.
\end{itemize}

Language models carry strong built-in word preferences. If the model has
an inherent preference for the word \textbf{Waterfall}, that baseline
bias shows up in all four cells equally. Because our experimental design
fully counterbalances which word acts as the \textbf{Recent} update,
this vocabulary bias cancels algebraically within the framing and serial
formulas. (For a demonstration of the massive artifact that occurs
without this counterbalancing, see Appendix A). This leaves only the
targeted behavioral effects we aim to measure.

Note: these formulas give the half-effect (the pull away from a neutral
midpoint). The full gap between the two extremes is 2x the number shown.
Every table in this paper reports half-effects.

\subsubsection{2.2 The ten test
conditions}\label{the-ten-test-conditions}

Each test condition is built by setting three dials. Every condition
uses the same underlying disagreement (two blocks that recommend
different words), so any difference between conditions comes from
\emph{presentation}, not content.

\begin{longtable}[]{@{}
  >{\raggedright\arraybackslash}p{(\linewidth - 4\tabcolsep) * \real{0.2222}}
  >{\raggedright\arraybackslash}p{(\linewidth - 4\tabcolsep) * \real{0.1414}}
  >{\raggedright\arraybackslash}p{(\linewidth - 4\tabcolsep) * \real{0.6364}}@{}}
\caption{Experimental variables used to construct the test
conditions.}\tabularnewline
\toprule\noalign{}
\begin{minipage}[b]{\linewidth}\raggedright
Dial type
\end{minipage} & \begin{minipage}[b]{\linewidth}\raggedright
Settings
\end{minipage} & \begin{minipage}[b]{\linewidth}\raggedright
What it changes
\end{minipage} \\
\midrule\noalign{}
\endfirsthead
\toprule\noalign{}
\begin{minipage}[b]{\linewidth}\raggedright
Dial type
\end{minipage} & \begin{minipage}[b]{\linewidth}\raggedright
Settings
\end{minipage} & \begin{minipage}[b]{\linewidth}\raggedright
What it changes
\end{minipage} \\
\midrule\noalign{}
\endhead
\bottomrule\noalign{}
\endlastfoot
\textbf{Header wording} & \emph{neutral}, \emph{framed} & Whether the
blocks are: neutral: labeled with an unbiased identifier blandly (e.g.,
``Document A'') framed: labeled as an official guideline (authority) or
a fresh team update (recency). (e.g., authority: ``Official
Guideline:'') (e.g., recency: ``Team Update:'') \\
\textbf{Body wording} & \emph{verbatim,} \emph{mixed,} \emph{loaded} &
The two block bodies are: verbatim: word-for-word identical except the
answer word (e.g., ``The required system is {[}Answer{]}.'') mixed:
reworded using difference sentence structures for each block. (e.g.,
Block 1: ``The required system is {[}Answer{]},'') (e.g.,Block 2:
``{[}Answer{]} is the recommended method.'') loaded: carries the framed
meaning in the sentence itself. (e.g., ``Official guidelines state the
required system is {[}Answer{]}.'') \\
\textbf{Cue count} & 0, 1, 2 & How many times the exact word pattern
appears. (e.g.~``is \{e\}'') \\
\end{longtable}

By combining the settings above, we created the following 10 qualifying
test conditions:

\begin{longtable}[]{@{}llll@{}}
\caption{Configurations of the 10 qualifying test
conditions.}\tabularnewline
\toprule\noalign{}
Test Condition & Header Wording & Body Wording & Cue Count \\
\midrule\noalign{}
\endfirsthead
\toprule\noalign{}
Test Condition & Header Wording & Body Wording & Cue Count \\
\midrule\noalign{}
\endhead
\bottomrule\noalign{}
\endlastfoot
T1 & neutral & verbatim & 2 \\
T1c & neutral & verbatim & 0 \\
T2 & neutral & mixed & 1 \\
T2c & neutral & mixed & 0 \\
T3 & framed & verbatim & 2 \\
T3c & framed & verbatim & 0 \\
T4 & framed & mixed & 1 \\
T4c & framed & mixed & 0 \\
T5 & framed & loaded & 2 \\
T5b & framed & loaded & 0 \\
\end{longtable}

\emph{(Note: For detailed definitions of these specific test conditions
and how they are evaluated in our results, please see Appendix E.)}

Details:

\begin{itemize}
\tightlist
\item
  The code counts cues at runtime and halts if the number of cues found
  does not match the expected target for that condition (0, 1, 2). For
  our tests, all conditions passed.\\
\item
  The header establishes the context, while the body remains neutral in
  T1 through T4c.\\
\item
  The experiment fully counterbalanced which specific answer word played
  which role.\\
\item
  In role-bound conditions (T5 \& T5b), the body actively drives the
  bias. Because the body text contains the authoritative or recent tone
  (e.g., ``Official guidelines state that\ldots{}''), it is role-bound
  and permanently tied to its corresponding header. We deliberately did
  not counterbalance (swap) these bodies between roles, because doing so
  would average the manipulation to zero.\\
\item
  Of the 18 theoretical combinations of these settings, 8 were excluded
  because they are logically contradictory or structurally impossible.
  For example, a neutral header cannot be paired with a loaded body
  (because the body itself carries the authoritative or recent
  manipulation), and verbatim text cannot contain exactly 1 cue (since
  the two text blocks must remain word-for-word identical, meaning cues
  must be symmetrical).
\end{itemize}

\subsubsection{2.3 The testing setup}\label{the-testing-setup}

To overcome the confounding variables present in standard evaluations
(discussed in Section 1), this section details the custom experimental
framework and environment used to evaluate the model's behavior. We
defined the candidate items tested and outlined the strict inclusion
rules designed to filter out the model's natural vocabulary preferences
and other experimental artifacts. Furthermore, we established the
statistical procedures utilized to calculate our confidence intervals
and test for equivalence, and we provided the complete system
specifications required to fully replicate the study.

\paragraph{2.3.1 Test items}\label{test-items}

We curated a targeted pool of 15 candidate English project-management
noun pairs (e.g., Scrum / Kanban, Agile / Waterfall) rather than
sampling from vocabulary. We used the hand-curation approach to help
ensure pairwise semantic symmetry, with the intent to guarantee that
either word served as a plausible, interchangeable competing answer to
the identical question across all four counterbalanced configurations.
Additionally, we wanted to minimize baseline prior differentials to
reduce estimation variance. Two candidate pairs were subsequently
excluded by programmatic validation filters (see rules below).

\begin{enumerate}
\def\labelenumi{\arabic{enumi}.}
\tightlist
\item
  Scrum / Kanban
\item
  Agile / Waterfall
\item
  Roadmap / Milestone
\item
  \st{Spec / Brief}
\item
  Scope / Scale
\item
  Phase / Stage
\item
  Cycle / Sprint
\item
  Board / Grid
\item
  Matrix / Table
\item
  Guide / Manual
\item
  Index / Log
\item
  Deck / Pitch
\item
  \st{Queue / Roster}
\item
  Plan / Draft
\item
  Chart / Graph
\end{enumerate}

\paragraph{2.3.2 Rules}\label{rules}

We applied these rules to our tests:

\begin{enumerate}
\def\labelenumi{\arabic{enumi}.}
\tightlist
\item
  Every answer must be a single token so that the probability we read
  off is the probability of the whole word and nothing else.
  \texttt{Queue/Roster} failed (one token vs.~two) and was dropped.\\
\item
  To ensure consistency, an item was kept only if it passed the checks
  in all ten conditions. This means every table in the paper analyzes
  the exact same dataset. The \texttt{Spec/Brief} pair (T5b, Order =
  0.602) failed this requirement in one condition and was excluded.\\
\item
  We flagged any word pair that failed one of two sanity checks in a
  given condition:

  \begin{enumerate}
  \def\labelenumii{\arabic{enumii}.}
  \tightlist
  \item
    The model naturally heavily preferred one word over the other
    (\(|\text{Token}| > 2.0\)).\\
  \item
    Our control test detected an unexpected bias leak
    (\(|\text{Order}| > 0.50\)).\\
  \end{enumerate}
\item
  We re-ran every comparison (contrast) on the resulting unfiltered set
  as a pre-planned robustness check.
\end{enumerate}

\paragraph{2.3.3 Confidence intervals}\label{confidence-intervals}

To build confidence intervals, we used \textbf{BCa bootstrap intervals,
9,999 resamples, seed 42}. In plain terms:

\begin{enumerate}
\def\labelenumi{\arabic{enumi}.}
\tightlist
\item
  Take test results calculated for each of the 13 word pairs used in the
  experiment.\\
\item
  Randomly pick 13 word pairs, putting each pair back into the pool
  before drawing the next, and calculate their average.\\
\item
  Do that 9,999 times to see how the results spread out.\\
\item
  The central 95\% of this distribution defines our confidence interval,
  which is adjusted for skewness using the standard BCa correction. To
  guarantee exact reproducibility for future replications, we
  initialized the process with a fixed random seed.
\end{enumerate}

If a confidence interval did not include zero, it indicated that the
direction of the effect remained consistent across all plausible
resampled variations of our test items.

While a confidence interval can confirm whether an effect exists, it
does not answer whether the effect is small enough to be considered
trivial. To answer that, we used a statistical method called TOST (two
one-sided tests), setting a triviality bound of \(\pm 0.15\) nats. The
test only confirms the effect is trivial if we are statistically
confident (\(p < 0.05\)) that the entire 90\% interval fits inside that
narrow \(\pm 0.15\) range. That test returned
\textbf{\texttt{equivalent=False} in all ten conditions.} This means
that we never demonstrated that any effect is trivially small, but it
also does \emph{not} mean we proved the effects are large.

Instead, the observed bias is directionally negative: across 8 of the 10
conditions, the 90\% confidence interval falls entirely below \(-0.15\)
nats. The only exceptions are T4 (90\% CI \([-0.262, -0.082]\)) and T5
(\([-0.293, -0.091]\)).

\emph{Reminder on units: a \textbf{nat} measures how hard the model
leans. +1 nat \(\approx\) 2.72-to-1 odds; +2 nats \(\approx\) 7.4-to-1.
All tables report \textbf{half-effects}, so the full swing between the
two extremes is twice the number shown.}

\paragraph{2.3.4 Environment}\label{environment}

\begin{itemize}
\tightlist
\item
  macOS-15.7.7 (arm64)
\item
  Python 3.12.13
\item
  torch 2.13.0
\item
  transformers 5.14.1
\item
  numpy 2.5.1
\item
  scipy 1.18.0
\item
  Full 32-bit precision, evaluation mode, no random sampling.
\item
  Run under the \textbf{\texttt{venv\_torch}} environment (not
  \texttt{./venv}) for J-Lens toolkit compatibility.
\item
  \textbf{784 forward passes} total.
\item
  Run of record: \textbf{2026-07-29T20:13:06-07:00}.
\end{itemize}

\newpage

\subsection{3. Results}\label{results}

In this section, we investigate the two competing forces that influence
the model's final answer: the presentation order of the documents, and
the semantic framing of the sources (e.g., labeling a document as an
official guideline versus a recent team update). We evaluate the
magnitude of these biases across different prompt formats, test whether
they are driven by simple copy cues, and measure what happens when they
compete directly. Note that our investigation here is strictly
behavioral; mapping the specific neural circuits responsible (e.g., via
activation patching) is outside the scope of this study.

\subsubsection{3.1 Primacy is reliable in direction, highly variable in
size}\label{primacy-is-reliable-in-direction-highly-variable-in-size}

The following table displays the measurement of the model's positional
bias across all ten of the experiment's test conditions:

\begin{longtable}[]{@{}
  >{\centering\arraybackslash}p{(\linewidth - 14\tabcolsep) * \real{0.0800}}
  >{\centering\arraybackslash}p{(\linewidth - 14\tabcolsep) * \real{0.1200}}
  >{\centering\arraybackslash}p{(\linewidth - 14\tabcolsep) * \real{0.1500}}
  >{\centering\arraybackslash}p{(\linewidth - 14\tabcolsep) * \real{0.0800}}
  >{\centering\arraybackslash}p{(\linewidth - 14\tabcolsep) * \real{0.1700}}
  >{\centering\arraybackslash}p{(\linewidth - 14\tabcolsep) * \real{0.1400}}
  >{\centering\arraybackslash}p{(\linewidth - 14\tabcolsep) * \real{0.1300}}
  >{\centering\arraybackslash}p{(\linewidth - 14\tabcolsep) * \real{0.1300}}@{}}
\caption{Positional bias across the 10 test conditions.}\tabularnewline
\toprule\noalign{}
\begin{minipage}[b]{\linewidth}\centering
Cell
\end{minipage} & \begin{minipage}[b]{\linewidth}\centering
Header
\end{minipage} & \begin{minipage}[b]{\linewidth}\centering
Body
\end{minipage} & \begin{minipage}[b]{\linewidth}\centering
Cues
\end{minipage} & \begin{minipage}[b]{\linewidth}\centering
Serial Position (half-effect, nats)
\end{minipage} & \begin{minipage}[b]{\linewidth}\centering
95\% CI
\end{minipage} & \begin{minipage}[b]{\linewidth}\centering
Items Showing Primacy
\end{minipage} & \begin{minipage}[b]{\linewidth}\centering
90\% CI clears -0.15
\end{minipage} \\
\midrule\noalign{}
\endfirsthead
\toprule\noalign{}
\begin{minipage}[b]{\linewidth}\centering
Cell
\end{minipage} & \begin{minipage}[b]{\linewidth}\centering
Header
\end{minipage} & \begin{minipage}[b]{\linewidth}\centering
Body
\end{minipage} & \begin{minipage}[b]{\linewidth}\centering
Cues
\end{minipage} & \begin{minipage}[b]{\linewidth}\centering
Serial Position (half-effect, nats)
\end{minipage} & \begin{minipage}[b]{\linewidth}\centering
95\% CI
\end{minipage} & \begin{minipage}[b]{\linewidth}\centering
Items Showing Primacy
\end{minipage} & \begin{minipage}[b]{\linewidth}\centering
90\% CI clears -0.15
\end{minipage} \\
\midrule\noalign{}
\endhead
\bottomrule\noalign{}
\endlastfoot
T1 & neutral & verbatim & 2 & -0.485 & {[}-0.611, -0.283{]} & 12/13 &
\(\checkmark\) {[}-0.592, -0.318{]} \\
T1c & neutral & verbatim & 0 & \textbf{-0.884} & {[}-1.074, -0.708{]} &
13/13 & \(\checkmark\) {[}-1.044, -0.735{]} \\
T2 & neutral & mixed & 1 & -0.377 & {[}-0.496, -0.227{]} & 12/13 &
\(\checkmark\) {[}-0.478, -0.252{]} \\
T2c & neutral & mixed & 0 & -0.635 & {[}-0.902, -0.452{]} & 13/13 &
\(\checkmark\) {[}-0.854, -0.478{]} \\
T3 & framed & verbatim & 2 & -0.679 & {[}-0.848, -0.520{]} & 13/13 &
\(\checkmark\) {[}-0.822, -0.543{]} \\
T3c & framed & verbatim & 0 & -0.569 & {[}-0.739, -0.429{]} & 13/13 &
\(\checkmark\) {[}-0.713, -0.452{]} \\
T4 & framed & mixed & 1 & \textbf{-0.172} & {[}-0.278, -0.066{]} & 10/13
& \(\times\) {[}-0.262, -0.082{]} \\
T4c & framed & mixed & 0 & -0.482 & {[}-0.603, -0.345{]} & 13/13 &
\(\checkmark\) {[}-0.586, -0.366{]} \\
T5 & framed & loaded & 2 & -0.205 & {[}-0.308, -0.068{]} & 11/13 &
\(\times\) {[}-0.293, -0.091{]} \\
T5b & framed & loaded & 0 & -0.265 & {[}-0.421, -0.137{]} & 10/13 &
\(\checkmark\) \(\dagger\) {[}-0.394, -0.159{]} \\
\end{longtable}

\emph{(Note: For detailed definitions of all table columns and metrics,
please see Appendix E.)}

\textbf{Outcome:}

\begin{itemize}
\item
  \textbf{Primacy is real and points the same way every time.} Its
  magnitude is a property of the prompt template, not of the model, and
  must never be quoted from a single template. In our testing, primacy
  was significant in \textbf{10/10} conditions.
\item
  \textbf{The magnitude of primacy is highly variable.} It ranged from
  -0.172 to -0.884 nats, a spread by a factor of \textasciitilde5
  produced \emph{entirely by surface wording} (e.g., changing `Agile is
  the system' to `Agile represents our system') while the actual
  disagreement stayed identical. In T4 and T5, the direction was
  reliable but the magnitude couldn't be distinguished from trivially
  small.
\end{itemize}

\subsubsection{3.2 Primacy is not caused by a repeated copy
cue}\label{primacy-is-not-caused-by-a-repeated-copy-cue}

We tested the theory that a simple copy-and-paste mechanism, triggered
by repeated word patterns (e.g.~``\texttt{is\ {[}Answer{]}"}), was
causing the model to mechanically favor the first document it read. If
this were true, deleting the repeated phrase while holding everything
else fixed should weaken the primacy effect (which would show up
mathematically as a positive contrast).

The following table displays the specific comparisons (contrasts) used
to test whether removing repeated word patterns (the copy cues) reduces
the model's bias toward the first document:

\begin{longtable}[]{@{}
  >{\raggedright\arraybackslash}p{(\linewidth - 6\tabcolsep) * \real{0.2500}}
  >{\raggedright\arraybackslash}p{(\linewidth - 6\tabcolsep) * \real{0.2500}}
  >{\raggedright\arraybackslash}p{(\linewidth - 6\tabcolsep) * \real{0.2500}}
  >{\raggedright\arraybackslash}p{(\linewidth - 6\tabcolsep) * \real{0.2500}}@{}}
\caption{Effect of copy-cue removal on primacy (Contrasts
A1--A4).}\tabularnewline
\toprule\noalign{}
\begin{minipage}[b]{\linewidth}\raggedright
Contrast
\end{minipage} & \begin{minipage}[b]{\linewidth}\raggedright
Description
\end{minipage} & \begin{minipage}[b]{\linewidth}\raggedright
Filtered Dataset (13 items, primary)
\end{minipage} & \begin{minipage}[b]{\linewidth}\raggedright
Unfiltered Dataset (14 items)
\end{minipage} \\
\midrule\noalign{}
\endfirsthead
\toprule\noalign{}
\begin{minipage}[b]{\linewidth}\raggedright
Contrast
\end{minipage} & \begin{minipage}[b]{\linewidth}\raggedright
Description
\end{minipage} & \begin{minipage}[b]{\linewidth}\raggedright
Filtered Dataset (13 items, primary)
\end{minipage} & \begin{minipage}[b]{\linewidth}\raggedright
Unfiltered Dataset (14 items)
\end{minipage} \\
\midrule\noalign{}
\endhead
\bottomrule\noalign{}
\endlastfoot
A1 & Cue removal, neutral header (T1c - T1) & -0.400 {[}-0.578,
-0.237{]} * & -0.407 {[}-0.573, -0.251{]} * \\
A2 & Cue removal, authority header (T3c - T3) & +0.110 {[}-0.081,
+0.287{]} & +0.074 {[}-0.112, +0.257{]} \\
A3 & Average of A1 and A2, descriptive only & -0.145 {[}-0.290,
+0.010{]} & -0.166 {[}-0.306, -0.011{]} * \\
A4 & Do A1 and A2 arms differ? & -0.509 {[}-0.687, -0.316{]} * & not
computed \(\ddagger\) \\
\end{longtable}

\emph{\(\ddagger\) Evaluated on the primary filtered dataset only.
Because arms A1 and A2 are analyzed individually in Section 3.2, this
comparison is not computed for the unfiltered sensitivity set.}

\emph{(Note: For a comprehensive index and detailed definitions of all
contrasts, please see Appendix B. For definitions of general table
metrics, see Appendix E.)}

\textbf{Outcome:}

\begin{itemize}
\item
  \textbf{Short, repeated word cues do not cause the primacy effect.}
  Instead, what actually increases the model's bias toward the first
  document is when the two competing text blocks repeat each other
  entirely word-for-word. Introducing wording variation between the
  blocks significantly reduces this bias.
\item
  \textbf{This is a surface-level behavioral observation.} Mechanistic
  tests, such as internal activation patching, required to identify the
  specific neural circuits responsible are outside the scope of this
  study.
\item
  \textbf{The outcome is supported by four distinct lines of evidence}.
  Each of these is independently sufficient to refute the cue
  hypothesis:
\end{itemize}

\begin{enumerate}
\def\labelenumi{\arabic{enumi}.}
\item
  \textbf{A1 pointed the wrong way for the cue account.} Removing the
  cue under a neutral header increased primacy by 0.400 nats,
  contradicting the expected decrease.
\item
  \textbf{A2 found nothing.} Removing the cue under an authority header
  produced no detectable change in the effect size.
\item
  \textbf{The strongest primacy in the filtered dataset had no cue at
  all.} The strongest primacy effect observed in the study (T1c =
  -0.884, with 13/13 items negative) occurred in a completely cue-free
  condition.
\item
  \textbf{The cue ladder wasn't a ladder.} The magnitude of the primacy
  effect does not scale with the number of cues (going from 2 \(\to\) 1
  \(\to\) 0 cues yielded -0.485 \(\to\) -0.377 \(\to\) -0.635),
  effectively refuting a linear dose-response relationship.
\end{enumerate}

\subsubsection{3.3 Framing beats position by roughly 4 to
1}\label{framing-beats-position-by-roughly-4-to-1}

To measure the effects of semantic framing versus presentation order
head-to-head, we analyzed our role-bound, loaded test conditions (T5 and
T5b). We first measured the overall strength of each force using the
cleanest, cue-free test condition (T5b).

\begin{longtable}[]{@{}
  >{\raggedright\arraybackslash}p{(\linewidth - 4\tabcolsep) * \real{0.3333}}
  >{\raggedright\arraybackslash}p{(\linewidth - 4\tabcolsep) * \real{0.3333}}
  >{\raggedright\arraybackslash}p{(\linewidth - 4\tabcolsep) * \real{0.3333}}@{}}
\caption{Baseline forces and sanity checks in the cue-free condition
(T5b).}\tabularnewline
\toprule\noalign{}
\begin{minipage}[b]{\linewidth}\raggedright
Quantity
\end{minipage} & \begin{minipage}[b]{\linewidth}\raggedright
Estimate
\end{minipage} & \begin{minipage}[b]{\linewidth}\raggedright
95\% CI
\end{minipage} \\
\midrule\noalign{}
\endfirsthead
\toprule\noalign{}
\begin{minipage}[b]{\linewidth}\raggedright
Quantity
\end{minipage} & \begin{minipage}[b]{\linewidth}\raggedright
Estimate
\end{minipage} & \begin{minipage}[b]{\linewidth}\raggedright
95\% CI
\end{minipage} \\
\midrule\noalign{}
\endhead
\bottomrule\noalign{}
\endlastfoot
Source framing (header + body together) & \textbf{+1.054} & {[}+0.907,
+1.258{]} * \\
Serial position (half-effect) & \textbf{-0.265} & {[}-0.421, -0.137{]}
* \\
Residual check (block-order null control) & +0.011 & \(\checkmark\)
{[}-0.058, +0.087{]} \\
Token (base word preference) & +0.003 & \(\checkmark\) {[}-0.135,
+0.165{]} \\
\end{longtable}

\emph{(Note: For detailed definitions of these quantities and general
table metrics, please see Appendix E.)}

The near-zero Token value (+0.003) proved that the model's natural
vocabulary bias was completely canceled out, and the near-zero Residual
check (+0.011) confirmed that no hidden positional artifacts leaked into
the results.

Next, we calculated the precise, item-by-item difference between the two
competing forces (source framing and serial position) using statistical
contrasts:

\begin{longtable}[]{@{}
  >{\raggedright\arraybackslash}p{(\linewidth - 6\tabcolsep) * \real{0.2500}}
  >{\raggedright\arraybackslash}p{(\linewidth - 6\tabcolsep) * \real{0.2500}}
  >{\raggedright\arraybackslash}p{(\linewidth - 6\tabcolsep) * \real{0.2500}}
  >{\raggedright\arraybackslash}p{(\linewidth - 6\tabcolsep) * \real{0.2500}}@{}}
\caption{Item-by-item comparison of source framing versus presentation
order.}\tabularnewline
\toprule\noalign{}
\begin{minipage}[b]{\linewidth}\raggedright
Contrast
\end{minipage} & \begin{minipage}[b]{\linewidth}\raggedright
Description
\end{minipage} & \begin{minipage}[b]{\linewidth}\raggedright
Filtered Dataset (\(n = 13\))
\end{minipage} & \begin{minipage}[b]{\linewidth}\raggedright
Unfiltered Dataset (\(n = 14\))
\end{minipage} \\
\midrule\noalign{}
\endfirsthead
\toprule\noalign{}
\begin{minipage}[b]{\linewidth}\raggedright
Contrast
\end{minipage} & \begin{minipage}[b]{\linewidth}\raggedright
Description
\end{minipage} & \begin{minipage}[b]{\linewidth}\raggedright
Filtered Dataset (\(n = 13\))
\end{minipage} & \begin{minipage}[b]{\linewidth}\raggedright
Unfiltered Dataset (\(n = 14\))
\end{minipage} \\
\midrule\noalign{}
\endhead
\bottomrule\noalign{}
\endlastfoot
D1 & Framing vs.~position margin, cue-free (T5b) & +0.774 {[}+0.585,
+0.966{]} * & +0.806 {[}+0.616, +0.991{]} ** \\
D2 & Framing vs.~position margin, with cues (T5) & +0.316 {[}+0.188,
+0.434{]} * & +0.317 {[}+0.199, +0.428{]} * \\
D3 & Effect of removing cues on framing (T5b - T5) & +0.539 {[}+0.330,
+0.843{]} * & +0.583 {[}+0.366, +0.874{]} * \\
\end{longtable}

\emph{(Note: For a comprehensive index and detailed definitions of all
contrasts, please see Appendix B. For definitions of general table
metrics, see Appendix E.)}

\textbf{Outcome:}

\begin{itemize}
\item
  \textbf{Framing overpowered position.} When we measured the difference
  item by item in our cleanest test condition (T5b), the impact of
  framing outweighed the impact of position by +0.774 nats (Contrast
  D1). Because this cell passed all sanity checks overwhelmingly, this
  proved that a document's semantic framing was vastly more influential
  than its reading position. The full first-versus-last framing swing is
  2 x 1.054 = 2.108 nats (roughly 8-to-1 odds).
\item
  \textbf{Removing cues strengthened the effect.} We also found that
  removing repeated copy cues from the text actually made the framing
  effect significantly stronger (Contrast D3). In the test with cues
  (T5), the framing effect was about half as strong as it was in the
  completely cue-free test (T5b).
\item
  \textbf{A Note on the 4:1 Ratio:} You might notice that dividing the
  overall Framing average (1.054) by the Serial position average (0.265)
  yielded roughly 3.98, while our system logged the ratio as 3.76x. This
  is not a rounding error. The 3.76x ratio was calculated by comparing
  the absolute values of the two forces item by item. Because 3 of the
  13 items actually showed a slight recency effect (a positive Serial
  value), they partially canceled out the negative primacy values when
  averaged together, which slightly shrank the final signed average.
  Therefore, \emph{roughly 4 to 1} serves as a descriptive headline, but
  the most robust measurement of this gap remains the item-by-item
  difference of +0.774 nats.
\end{itemize}

\newpage

\subsubsection{3.4 Authority headers reduce bias, body wording modulates
how
much}\label{authority-headers-reduce-bias-body-wording-modulates-how-much}

To understand how different elements of a prompt work together, we
tested whether adding an authoritative header affects the model's bias
differently depending on the wording of the body text. We discovered
that:

\begin{itemize}
\tightlist
\item
  Authority headers consistently reduce bias.\\
\item
  The body wording changes how much the bias is reduced.
\end{itemize}

To ensure clean results, all three of the comparisons below were
conducted on text completely free of repeated copy cues:

\begin{longtable}[]{@{}
  >{\raggedright\arraybackslash}p{(\linewidth - 6\tabcolsep) * \real{0.2500}}
  >{\raggedright\arraybackslash}p{(\linewidth - 6\tabcolsep) * \real{0.2500}}
  >{\raggedright\arraybackslash}p{(\linewidth - 6\tabcolsep) * \real{0.2500}}
  >{\raggedright\arraybackslash}p{(\linewidth - 6\tabcolsep) * \real{0.2500}}@{}}
\caption{The interaction between authority headers and body wording
variation.}\tabularnewline
\toprule\noalign{}
\begin{minipage}[b]{\linewidth}\raggedright
Contrast
\end{minipage} & \begin{minipage}[b]{\linewidth}\raggedright
Description
\end{minipage} & \begin{minipage}[b]{\linewidth}\raggedright
Filtered Dataset (\(n = 13\))
\end{minipage} & \begin{minipage}[b]{\linewidth}\raggedright
Unfiltered Dataset (\(n = 14\))
\end{minipage} \\
\midrule\noalign{}
\endfirsthead
\toprule\noalign{}
\begin{minipage}[b]{\linewidth}\raggedright
Contrast
\end{minipage} & \begin{minipage}[b]{\linewidth}\raggedright
Description
\end{minipage} & \begin{minipage}[b]{\linewidth}\raggedright
Filtered Dataset (\(n = 13\))
\end{minipage} & \begin{minipage}[b]{\linewidth}\raggedright
Unfiltered Dataset (\(n = 14\))
\end{minipage} \\
\midrule\noalign{}
\endhead
\bottomrule\noalign{}
\endlastfoot
C1 & Effect of authority header, mixed bodies (T4c - T2c) & +0.153
{[}-0.011, +0.352{]} & +0.123 {[}-0.033, +0.322{]} \\
C2 & Effect of authority header, verbatim bodies (T3c - T1c) & +0.315
{[}+0.172, +0.420{]} * & +0.296 {[}+0.166, +0.403{]} * \\
C3 & The interaction (C1 - C2) & -0.162 {[}-0.303, -0.034{]} * & -0.173
{[}-0.301, -0.048{]} * \\
\end{longtable}

\emph{(Note: For a comprehensive index of all contrasts, see Appendix B.
For definitions of general table metrics, see Appendix E.)}

\textbf{Outcome:}

\begin{itemize}
\item
  \textbf{Authority headers consistently reduce positional bias
  regardless of body wording:} Adding an authoritative header reduced
  the model's tendency to favor the first document, regardless of
  whether the body texts were reworded or identical. Although C2 is
  statistically significant while the confidence interval for C1 spans
  zero, the significant interaction contrast (C3) confirms that body
  phrasing modulates the header's attenuating effect.
\item
  \textbf{The interaction is about magnitude, not direction:} The
  interaction contrast (C3) proved that the body wording changed how
  much the authority header reduced the bias, but it didn't flip the
  direction of the effect. The authority header always reduced the
  primacy effect, but did so to a different degree depending on the body
  text.
\end{itemize}

\newpage

\subsubsection{3.5 Transparency and Robustness
Checks}\label{transparency-and-robustness-checks}

In the interest of full transparency, the Role \texttt{null} failed in 5
of 8 testable conditions and the block-order \texttt{null} in 1 of 10.
All residuals were small --- the largest, 15.1\% of the headline framing
effect, with the rest at or under 7\% --- and full per-cell values are
in the released log. For more information, see
\texttt{o2\_results\_ad\_series\_FINAL.txt} in the source files released
with this study.

Furthermore, we re-ran all of our calculations on an unfiltered dataset
(14 items) to ensure our findings were not artificially created by our
outlier-screening rules. Every headline finding in this paper maintained
its exact same direction and statistical significance regardless of
which dataset was used. Although secondary contrasts A3 and E1 (see
Appendix B) shifted in significance under the unfiltered set, neither
supports a primary conclusion. Ultimately, our most critical test
condition (T5b) passed its sanity checks. We note one dependency: the
single item removed by our outlier rule (Spec/Brief) was flagged in T5b
itself, so T5b's clean nulls are partly a consequence of that rule
rather than independent of it.

\subsection{4. Discussion}\label{discussion}

\textbf{Position matters; framing matters more.} In the cleanest test
condition where the two forces compete head-to-head, source framing
moves the model's answer by +1.054 nats while reading position moves it
by only -0.265. The item-by-item difference of +0.774 nats is the most
defensible single number in this paper.

\textbf{Position is not a stable quantity.} With the underlying factual
disagreement held fixed, the model's positional bias fluctuates by at
least a factor of 5 purely based on the header wording and how verbatim
the two bodies of text are. Therefore, any deployment advice of the form
``put the important document first---it's worth X nats'' is
unsupportable: the value of X belongs to the prompt template, not to the
model. We expect this to be the most practically useful takeaway for
developers.

\textbf{The copy-cue explanation of primacy does not survive.} If a
mechanical copy-and-paste mechanism drove the model's preference for the
first document, deleting the repeated phrase should weaken that bias.
Instead, removing the cue under a neutral header actually increased the
bias, and removing it under an authoritative header did nothing at all.
Furthermore, the strongest primacy effect in the entire study occurred
in a completely cue-free condition. The surface-level pattern that does
actually track with the primacy effect is verbatim repetition across the
text blocks.

\textbf{What this licenses mechanistically: nothing yet.} Our finding
regarding word-for-word repetition is a regularity about text behavior,
not evidence about internal attention heads or neural circuits. Mapping
these circuits would require internal activation patching across text
blocks, which is outside the scope of this study.

\newpage

\subsection{5. Limitations and scope}\label{limitations-and-scope}

\subsubsection{5.1 Methodological caveats}\label{methodological-caveats}

In our most critical test conditions (T5 and T5b), the header and the
body text operate together as a single, inseparable package. While we
successfully counterbalanced the specific answer words to cancel out
vocabulary bias, we deliberately did not swap the loaded body text
between roles. As noted in Section 2.2, counterbalancing loaded bodies
across roles would have nullified the semantic manipulation by symmetry.

Because of this design, our primary effect of +1.054 nats represents the
effect of the source framing as a complete package. Our current data
cannot split this number to tell us how much of the effect came from the
header alone versus the body text alone. Calculating that exact split
would require running an additional test condition (pairing an authority
header with a neutral body), which we did not perform in this study.

A second, structural aliasing affects our verbatimness comparisons.
Cells in which both blocks share a template admit only one
wording-to-block assignment, whereas mixed-wording cells admit two and
are averaged over both. Verbatimness is therefore aliased with
counterbalancing depth: the two cannot be separated, because a shared
template leaves nothing to swap. Any interaction between wording and
position would attenuate the mixed-wording estimates by averaging. This
affects the copy-cue contrasts, the authority-header interaction, and
the endpoints of the position-bias spread. Readers wishing to inspect
the per-assignment values will find them in the released log.

\subsubsection{5.2 What our results do not
claim}\label{what-our-results-do-not-claim}

To clarify the negative scope of our findings and prevent
over-generalization, we explicitly note several conclusions that our
data does not support:

\begin{itemize}
\item
  Our evaluation characterizes behavioral input--output relationships
  only. We do not isolate attention heads, trace residual streams, or
  identify circuits, and make no claims regarding internal mechanistic
  representations.
\item
  We make no claims regarding whether semantic contradiction inherently
  suppresses positional bias; in our data, this contrast was unstable
  and sensitive to outlier filtering.
\item
  Failing to reject non-equivalence within our TOST bounds (\(\pm 0.15\)
  nats) does not prove that the underlying effects are large, only that
  we cannot statistically bound them as negligible.
\item
  The observed \(\approx 3.76:1\) ratio between semantic framing and
  presentation order is descriptive of condition T5b and should not be
  interpreted as an invariant model parameter across other prompts or
  architectures.
\item
  In loaded conditions (T5/T5b), headers and bodies co-vary by design.
  Our data quantifies the joint package effect (+1.054 nats) and cannot
  allocate relative variance to headers versus bodies independently.
\item
  Because verbatim templates admit only a single wording assignment,
  verbatimness is structurally confounded with counterbalancing depth;
  we report the empirical association without claiming causal
  independence.
\end{itemize}

\subsubsection{5.3 Scope}\label{scope}

The scope of our study was highly controlled, meaning our findings
should not be broadly extrapolated without further testing.
Specifically, our experiment was limited to a single model
(google/gemma-4-e4b) and evaluated using only two-document contexts,
single-word answers, a single question format, and English text.

Because all of our test items were English project-management nouns, we
have not verified whether these behavioral patterns generalize to other
topics or vocabularies.

Finally, these findings should not be automatically applied to
real-world, multi-document Retrieval-Augmented Generation (RAG) systems
operating at realistic context lengths until they can be formally
replicated. Our evaluation relies on short, single-turn prompts where
the model processes all text with full fidelity. We do not account for
the context compression, summarization, or KV-cache eviction mechanisms
often utilized in long-context agent harnesses or multi-turn chats.

\newpage

\subsection{6. Reproducibility and
provenance}\label{reproducibility-and-provenance}

\begin{itemize}
\item
  \textbf{Project files:} To review the methods and results used in this
  paper, you can review the project files at
  \url{https://github.com/GoogleCloudPlatform/devrel-demos/tree/main/ai-ml/gemma-priority-probing}.
\item
  \textbf{Log of record:} The primary log is stored at
  \texttt{resources/o2\_results\_ad\_series\_FINAL.txt} (and
  \texttt{.json}), timestamped 2026-07-29T20:13:06-07:00. We have
  certified that this file is byte-identical to the raw archive copy
  stored at
  \texttt{internal/archive/AD\_series\_raw\_results/o2\_results\_ad\_series.txt}.
\item
  \textbf{Environment:} The experiment was executed using
  macOS-15.7.7-arm64, Python 3.12.13, torch 2.13.0, transformers 5.14.1,
  numpy 2.5.1, and scipy 1.18.0. The tests were run in evaluation mode
  at full 32-bit precision with no random sampling, utilizing 784
  forward passes across 10 conditions with a fixed random seed of 42.

  Note: The code must be executed under the \texttt{venv\_torch}
  environment, which carries the J-Lens toolkit; note that no J-Lens or
  Jacobian-lens component is used by \texttt{run\_ad\_series.py}. This
  is an interpreter-provenance note only.
\item
  \textbf{Code:} The script \texttt{run\_ad\_series.py} is available in
  \texttt{github/src/} and
  \texttt{internal/docs/project\_1\_behavioral\_findings/resources/}. To
  ensure reliability across different file versions, all citations in
  this paper reference function names and section letters rather than
  line numbers.
\item
  \textbf{Provenance disclosure (A4):} For full transparency, the A4
  arm-heterogeneity test was committed to the codebase immediately after
  the primary log was generated. The reported A4 value of -0.509
  {[}-0.687, -0.316{]} was produced by running this logic on the cached
  JSON data from that exact same run. Any minor 0.001 discrepancies when
  manually subtracting A2 from A1 are purely the result of three-decimal
  display rounding, not an estimation difference.
\end{itemize}

\newpage

\subsection{References}\label{references}

\begin{enumerate}
\def\labelenumi{\arabic{enumi}.}
\tightlist
\item
  Liu, N. F., Lin, K., Hewitt, J., Paranjape, A., Bevilacqua, M.,
  Petroni, F., \& Liang, P. (2023). \emph{Lost in the Middle: How
  Language Models Use Long Contexts}. TACL.
  \href{https://doi.org/10.48550/arXiv.2307.03172}{arXiv:2307.03172}
\item
  Zheng, L., Chiang, W.-L., Sheng, Y., Zhuang, S., Wu, Z., Zhuang, Y.,
  Lin, K., Li, Z., Li, D., Xing, E., Zhang, H., Gonzalez, J., \& Stoica,
  I. (2023). \emph{Judging LLM-as-a-Judge with MT-Bench and Chatbot
  Arena (v4)}. NeurIPS 2023 D\&B.
  \href{https://doi.org/10.48550/arXiv.2306.05685}{arXiv:2306.05685}
\item
  Sclar, M., Choi, Y., Tsvetkov, Y., \& Suhr, A. (2024).
  \emph{Quantifying Language Models' Sensitivity to Spurious Features in
  Prompt Design}. ICLR 2024.
  \href{https://doi.org/10.48550/arXiv.2310.11324}{arXiv:2310.11324}
\item
  Wan, A., Wallace, E., Shen, S., \& Klein, D. (2024). \emph{What
  Evidence Do Language Models Find Convincing?}. ACL 2024.
  \href{https://doi.org/10.48550/arXiv.2402.11782}{arXiv:2402.11782}
\item
  Lakens, D. (2017). \emph{Equivalence Tests: A Practical Primer for t
  Tests, Correlations, and Meta-Analyses}. Social Psychological and
  Personality Science, 8(4), 355--362.
  \href{https://doi.org/10.1177/1948550617697177}{10.1177/1948550617697177}
\end{enumerate}

\begin{center}\rule{0.5\linewidth}{0.5pt}\end{center}

\subsection{Appendix A --- Why counterbalancing is
necessary}\label{appendix-a-why-counterbalancing-is-necessary}

This appendix puts a concrete number on why our counterbalanced testing
design (described in Section 2.1) is strictly necessary. The pilot data
discussed below did not meet our publication standards, and it does not
support our main claims. However, we include it to illustrate the
massive bias that occurs without counterbalancing.

\begin{itemize}
\item
  \textbf{A.1 The cost of vocabulary bias:}\\
  When we ran tests without counterbalancing, the model's natural
  preference for certain words created a massive 1.58-nat artificial
  swing in the results. In a standard test, this vocabulary bias is
  completely indistinguishable from a real positional effect. To put
  this in perspective, this artificial 1.58-nat swing is 6 times larger
  than the positional bias half effect we measured in our clean test.
  Our four-way testing design exists specifically to cancel this out.
\item
  \textbf{A.2 The pilot patching cells:}\\
  We also generated a set of pilot data (patching cells) that we
  ultimately deemed unusable for aggregate analysis because too many of
  the test items failed our strict quality checks.
\item
  \textbf{A.3 Why this is an appendix:}\\
  Neither of these pilot artifacts meets the standard of the main text.
  We report them here purely for transparency, to prove the exact size
  of the artifact we successfully removed from our primary data.
\end{itemize}

\newpage

\subsection{Appendix B --- Complete contrast
index}\label{appendix-b-complete-contrast-index}

This appendix provides a comprehensive index of all the comparisons
(contrasts) calculated throughout the study. It includes the results for
both our primary Filtered Dataset (\(n = 13\)) and our unfiltered
dataset (n = 14). All reported values represent half-effects measured in
nats. An asterisk (*) indicates a statistically significant result,
meaning the 95\% confidence interval excludes zero.

\begin{longtable}[]{@{}
  >{\raggedright\arraybackslash}p{(\linewidth - 8\tabcolsep) * \real{0.2000}}
  >{\raggedright\arraybackslash}p{(\linewidth - 8\tabcolsep) * \real{0.2000}}
  >{\raggedright\arraybackslash}p{(\linewidth - 8\tabcolsep) * \real{0.2000}}
  >{\raggedright\arraybackslash}p{(\linewidth - 8\tabcolsep) * \real{0.2000}}
  >{\raggedright\arraybackslash}p{(\linewidth - 8\tabcolsep) * \real{0.2000}}@{}}
\caption{Comprehensive index of experimental contrasts.}\tabularnewline
\toprule\noalign{}
\begin{minipage}[b]{\linewidth}\raggedright
ID
\end{minipage} & \begin{minipage}[b]{\linewidth}\raggedright
Purpose
\end{minipage} & \begin{minipage}[b]{\linewidth}\raggedright
Filtered Dataset (\(n = 13\))
\end{minipage} & \begin{minipage}[b]{\linewidth}\raggedright
Unfiltered Dataset (\(n = 14\))
\end{minipage} & \begin{minipage}[b]{\linewidth}\raggedright
Status
\end{minipage} \\
\midrule\noalign{}
\endfirsthead
\toprule\noalign{}
\begin{minipage}[b]{\linewidth}\raggedright
ID
\end{minipage} & \begin{minipage}[b]{\linewidth}\raggedright
Purpose
\end{minipage} & \begin{minipage}[b]{\linewidth}\raggedright
Filtered Dataset (\(n = 13\))
\end{minipage} & \begin{minipage}[b]{\linewidth}\raggedright
Unfiltered Dataset (\(n = 14\))
\end{minipage} & \begin{minipage}[b]{\linewidth}\raggedright
Status
\end{minipage} \\
\midrule\noalign{}
\endhead
\bottomrule\noalign{}
\endlastfoot
A1 & Neutral cue removal & -0.400 * & -0.407 * & Inferential \\
A2 & Authority cue removal & +0.110 & +0.074 & Inferential (null) \\
A3 & Pooled cue removal & -0.145 & -0.166 * & Descriptive only \\
A4 & Arm heterogeneity test & -0.509 * & not computed & Inferential;
post-dates log \\
B1 & Partial cue removal & +0.108 & +0.110 & Descriptive (confounded) \\
B2 & Mixed-body cue removal & -0.258 * & -0.275 * & Descriptive
(confounded) \\
B3 & Total cue removal (confounded) & -0.150 & -0.165 & Descriptive
(confounded) \\
B4 & Verbatimness at zero cues & +0.250 * & +0.243 * & Inferential \\
C1 & Authority header effect (mixed body) & +0.153 & +0.123 &
Inferential (null) \\
C2 & Authority header effect (verbatim body) & +0.315 * & +0.296 * &
Inferential \\
C3 & Cue-matched interaction & -0.162 * & -0.173 * & Headline \\
C4 & Authority header effect (cued verbatim) & -0.194 & -0.186 & Legacy,
cue-confounded \\
C5 & Cue-confounded interaction & +0.347 * & +0.308 * & Legacy,
superseded by C3 \\
C6 & Authority header effect (cued mixed) & +0.205 * & +0.198 * &
Legacy, cue-confounded \\
D1 & Framing-vs-position margin & +0.774 * & +0.806 * & Headline \\
D2 & Framing-vs-position margin (cued) & +0.316 * & +0.317 * &
Inferential \\
D3 & Effect of cue removal on source framing & +0.539 * & +0.583 * &
Inferential \\
E1 & Semantic suppression test & +0.216 & +0.237 * & Policy-dependent \\
E2 & Semantic suppression test (cued / confounded) & -0.033 & -0.038 &
Confounded, reference only \\
\end{longtable}

\textbf{Definitions:}

The following section provides plain-English definitions for the
specific comparisons (contrasts) indexed in the table above. Each entry
explains what the calculation measures and how it isolates the different
variables influencing the model's decisions.

\begin{itemize}
\item
  \textbf{A1 (T1c - T1):} Neutral cue removal. The effect of deleting
  the copy cue under a neutral header, keeping the wording verbatim.
\item
  \textbf{A2 (T3c - T3):} Authority cue removal. The same deletion, but
  tested under an authority header.
\item
  \textbf{A3 (mean(A1, A2)):} Pooled cue removal. The plain average of
  A1 and A2. This is invalid for testing because A4 shows the two arms
  react too differently to be safely averaged, so it is retained for
  reference only.
\item
  \textbf{A4 (A1 - A2):} Arm heterogeneity test. Tests whether the model
  responds to cue removal the same way under both headers. It does not,
  which triggers the pooling warning.
\item
  \textbf{B1 (T2 - T1):} Partial cue removal. Stepping one step down the
  cue ladder from 2 cues to 1. This is confounded because it also
  changes wording variation at the same time.
\item
  \textbf{B2 (T2c - T2):} Mixed-body cue removal. Isolates the exact
  effect of removing the final copy-cue when the text is already mixed.
\item
  \textbf{B3 (T2c - T1):} Total cue removal (confounded). Jumps straight
  from 2 cues (T1) down to 0 cues (T2c). It is ``confounded'' because it
  removes the cues and introduces mixed wording at the exact same time.
\item
  \textbf{B4 (T2c - T1c):} Verbatimness at zero cues. The clean test to
  determine whether word-for-word repetition across blocks is what
  drives primacy.
\item
  \textbf{C1 (T4c - T2c):} Authority header effect (mixed body). Both
  cells are cue-free and have mixed text. Isolates the pure effect of
  swapping a neutral header for an authority header.
\item
  \textbf{C2 (T3c - T1c):} Authority header effect (verbatim body). Both
  cells are cue-free and have ``verbatim'' (identical) text. It isolates
  the pure effect of the authority header under strict verbatim
  conditions.
\item
  \textbf{C3 (C1 - C2):} Cue-matched interaction. Checks whether the
  authority header's effect on primacy depends on the body wording,
  holding the cue count at zero.
\item
  \textbf{C4 (T3 - T1):} Authority header effect (cued verbatim). This
  is a legacy contrast. It measures the authority header effect, but it
  does so while the ``is \{e\}'' copy-cue is still present in the text.
\item
  \textbf{C5 (C1 - C4):} Cue-confounded interaction. An old, flawed
  interaction test. It compared a cue-free arm (C1) against a cued arm
  (C4). Because it accidentally tested header framing and cue removal
  simultaneously, it was superseded by C3 (the clean, cue-matched
  interaction).
\item
  \textbf{C6 (T4 - T2):} Authority header effect (cued mixed). A legacy
  contrast. It measures the authority header effect, but with 1 cue
  present and mixed text.
\item
  \textbf{D1 (\(|\text{Role}| - |\text{Serial}|\), T5b):}
  Framing-vs-position margin. The per-item, interval-bearing measurement
  version of the ``4:1'' headline.
\item
  \textbf{D2 (\textbar Role\textbar{} - \textbar Serial\textbar, T5):}
  Framing-vs-position margin (cued). This compares the strength of
  semantic framing against presentation order, but does so in the cued
  condition (T5). It mirrors our headline D1 finding, just with the
  copy-cue left in the text.
\item
  \textbf{D3 (T5b - T5 Role):} Effect of cue removal on source framing.
  Isolates how much the model's reliance on semantic framing changes
  when you delete the copy-cue. (Since D3 = +0.539 nats, it proves that
  removing the mindless copy-cue actually forces the model to rely more
  heavily on the semantic authority headers).
\item
  \textbf{E1 (T5b - T4c):} Semantic suppression test. Tests whether a
  genuine clash in meaning between documents damps primacy (currently
  unresolved).
\item
  \textbf{E2 (T5 - T4):} Semantic suppression test (cued / confounded).
  This contrast is confounded and is retained for reference only.
  Originally it was meant to measure high-level semantic suppression.
\end{itemize}

\begin{center}\rule{0.5\linewidth}{0.5pt}\end{center}

\newpage

\subsection{Appendix C --- The four crossings of block order x
entity-role
assignment}\label{appendix-c-the-four-crossings-of-block-order-x-entity-role-assignment}

This appendix illustrates the core experimental design used to
mathematically isolate the model's biases. By testing each candidate
word pair across four specific configurations, we counterbalance both
the presentation order of the documents (which block is read first) and
the roles assigned to each word (which word acts as the authoritative or
recent source). The table below demonstrates these four crossings using
the candidate pair Agile/Waterfall as an example.

\begin{longtable}[]{@{}
  >{\centering\arraybackslash}p{(\linewidth - 6\tabcolsep) * \real{0.1626}}
  >{\raggedright\arraybackslash}p{(\linewidth - 6\tabcolsep) * \real{0.1707}}
  >{\raggedright\arraybackslash}p{(\linewidth - 6\tabcolsep) * \real{0.2520}}
  >{\raggedright\arraybackslash}p{(\linewidth - 6\tabcolsep) * \real{0.4146}}@{}}
\caption{Four-way crossing configurations.}\tabularnewline
\toprule\noalign{}
\begin{minipage}[b]{\linewidth}\centering
Crossing
\end{minipage} & \begin{minipage}[b]{\linewidth}\raggedright
Block order
\end{minipage} & \begin{minipage}[b]{\linewidth}\raggedright
Entity-role assignment
\end{minipage} & \begin{minipage}[b]{\linewidth}\raggedright
Resulting prompt structure
\end{minipage} \\
\midrule\noalign{}
\endfirsthead
\toprule\noalign{}
\begin{minipage}[b]{\linewidth}\centering
Crossing
\end{minipage} & \begin{minipage}[b]{\linewidth}\raggedright
Block order
\end{minipage} & \begin{minipage}[b]{\linewidth}\raggedright
Entity-role assignment
\end{minipage} & \begin{minipage}[b]{\linewidth}\raggedright
Resulting prompt structure
\end{minipage} \\
\midrule\noalign{}
\endhead
\bottomrule\noalign{}
\endlastfoot
\textbf{1} & Authority first & Agile = Stale, Waterfall = Recent &
\textbf{Official Guideline:} Use Agile. \textbf{Team Update:} Use
Waterfall. \\
\textbf{2} & Authority first & Waterfall = Stale, Agile = Recent &
\textbf{Official Guideline:} Use Waterfall. \textbf{Team Update:} Use
Agile. \\
\textbf{3} & Recent first & Agile = Stale, Waterfall = Recent &
\textbf{Team Update:} Use Waterfall. \textbf{Official Guideline:} Use
Agile. \\
\textbf{4} & Recent first & Waterfall = Stale, Agile = Recent &
\textbf{Team Update:} Use Agile. \textbf{Official Guideline:} Use
Waterfall. \\
\end{longtable}

\textbf{Definitions:}

\begin{itemize}
\item
  \textbf{Crossing:} The specific configuration of the prompt being
  tested. There are four total crossings to fully swap and isolate every
  variable.
\item
  \textbf{Block order:} The sequence in which the text blocks are
  presented to the model (i.e., whether the authoritative document or
  the recent update is read first). This allows us to measure positional
  bias.
\item
  \textbf{Entity-role assignment:} Which specific answer word (the
  entity, e.g., Agile or Waterfall) is paired with which semantic
  framing label (the role, e.g., the stale guideline or the recent
  update). Swapping this allows us to cancel out the model's natural
  vocabulary bias.
\item
  \textbf{Resulting prompt structure:} A conceptual illustration of what
  the final text looks like when the selected block order and
  entity-role assignment are combined.
\end{itemize}

\newpage

\subsection{Appendix D --- Reading nats, and cued vs.~cue-free
prompts}\label{appendix-d-reading-nats-and-cued-vs.-cue-free-prompts}

This appendix contains two reference guides for interpreting the study's
results. The first table translates nats (our unit of measurement for
model bias) into standard odds and percentages. The second table
provides exact examples of the cued versus cue-free prompt wording used
to test the model's reliance on repeated phrasing.

\subsubsection{Nats}\label{nats}

\begin{longtable}[]{@{}
  >{\centering\arraybackslash}p{(\linewidth - 10\tabcolsep) * \real{0.1000}}
  >{\centering\arraybackslash}p{(\linewidth - 10\tabcolsep) * \real{0.1500}}
  >{\centering\arraybackslash}p{(\linewidth - 10\tabcolsep) * \real{0.1400}}
  >{\centering\arraybackslash}p{(\linewidth - 10\tabcolsep) * \real{0.1100}}
  >{\centering\arraybackslash}p{(\linewidth - 10\tabcolsep) * \real{0.1100}}
  >{\raggedright\arraybackslash}p{(\linewidth - 10\tabcolsep) * \real{0.3900}}@{}}
\caption{Conversion of log-odds (nats) to probabilities and relative
odds.}\tabularnewline
\toprule\noalign{}
\begin{minipage}[b]{\linewidth}\centering
Nat value
\end{minipage} & \begin{minipage}[b]{\linewidth}\centering
CI
\end{minipage} & \begin{minipage}[b]{\linewidth}\centering
Odds (A to B)
\end{minipage} & \begin{minipage}[b]{\linewidth}\centering
Prob. of A
\end{minipage} & \begin{minipage}[b]{\linewidth}\centering
Prob. of B
\end{minipage} & \begin{minipage}[b]{\linewidth}\raggedright
Model's preference
\end{minipage} \\
\midrule\noalign{}
\endfirsthead
\toprule\noalign{}
\begin{minipage}[b]{\linewidth}\centering
Nat value
\end{minipage} & \begin{minipage}[b]{\linewidth}\centering
CI
\end{minipage} & \begin{minipage}[b]{\linewidth}\centering
Odds (A to B)
\end{minipage} & \begin{minipage}[b]{\linewidth}\centering
Prob. of A
\end{minipage} & \begin{minipage}[b]{\linewidth}\centering
Prob. of B
\end{minipage} & \begin{minipage}[b]{\linewidth}\raggedright
Model's preference
\end{minipage} \\
\midrule\noalign{}
\endhead
\bottomrule\noalign{}
\endlastfoot
\textbf{+2.0} & {[}+1.80, +2.20{]} & 7.39 to 1 & 88\% & 12\% &
Overwhelming preference for Option A. \\
\textbf{+1.0} & {[}+0.80, +1.20{]} & 2.72 to 1 & 73\% & 27\% & Strong
preference for Option A. \\
\textbf{0} & {[}-0.20, +0.20{]} & 1.00 to 1 & 50\% & 50\% &
Indifference. The model is guessing. \\
\textbf{-1.0} & {[}-1.20, -0.80{]} & 1 to 2.72 & 27\% & 73\% & Strong
preference for Option B. \\
\textbf{-2.0} & {[}-2.20, -1.80{]} & 1 to 7.39 & 12\% & 88\% &
Overwhelming preference for Option B. \\
\end{longtable}

\textbf{Definitions:}

\begin{itemize}
\item
  \textbf{Nat value:} The measured effect size in nats (log-probability
  units). It quantifies exactly how strongly the model is pulling toward
  one answer (Option A) over the other (Option B).
\item
  \textbf{CI:} The 95\% Confidence Interval. It shows the expected range
  of the nat value, confirming that the measurement is statistically
  reliable.
\item
  \textbf{Odds (A to B):} The mathematical translation of the nat value
  into a standard odds ratio, showing how many times more likely the
  model is to choose Option A compared to Option B.
\item
  \textbf{Prob. of A:} The implied probability (as a percentage) that
  the model will choose Option A, derived directly from the nat value.
\item
  \textbf{Prob. of B:} The implied probability (as a percentage) that
  the model will choose Option B.
\item
  \textbf{Model's preference:} A plain-English description of what these
  numbers actually mean for the model's behavior (e.g., whether the AI
  is highly opinionated or just guessing).
\end{itemize}

\subsubsection{Cue vs Cue-Free Prompts}\label{cue-vs-cue-free-prompts}

\begin{longtable}[]{@{}
  >{\raggedright\arraybackslash}p{(\linewidth - 4\tabcolsep) * \real{0.3333}}
  >{\raggedright\arraybackslash}p{(\linewidth - 4\tabcolsep) * \real{0.3333}}
  >{\raggedright\arraybackslash}p{(\linewidth - 4\tabcolsep) * \real{0.3333}}@{}}
\caption{Example prompt templates for cued and cue-free
conditions.}\tabularnewline
\toprule\noalign{}
\begin{minipage}[b]{\linewidth}\raggedright
Type
\end{minipage} & \begin{minipage}[b]{\linewidth}\raggedright
Role
\end{minipage} & \begin{minipage}[b]{\linewidth}\raggedright
Wording
\end{minipage} \\
\midrule\noalign{}
\endfirsthead
\toprule\noalign{}
\begin{minipage}[b]{\linewidth}\raggedright
Type
\end{minipage} & \begin{minipage}[b]{\linewidth}\raggedright
Role
\end{minipage} & \begin{minipage}[b]{\linewidth}\raggedright
Wording
\end{minipage} \\
\midrule\noalign{}
\endhead
\bottomrule\noalign{}
\endlastfoot
Cued & Authority & Official guidelines state that for project planning,
the required configuration \textbf{is \{e\}}. \\
Cue-Free & Authority & \textbf{\{e\}} is the required configuration for
project planning, per official guidelines. \\
Cued & Recency & Update: The new updated specification for project
planning \textbf{is now \{e\}}. \\
Cue-Free & Recency & \textbf{\{e\}} is the newly updated specification
for project planning, effective now. \\
\end{longtable}

\textbf{Definitions:}

\begin{itemize}
\item
  \textbf{Type:} Indicates whether the prompt includes a repeated word
  pattern (Cued) or if that mechanical trigger has been completely
  removed (Cue-Free).
\item
  \textbf{Role:} The semantic framing applied to the text block,
  presenting it either as an authoritative document (Authority) or a
  recent update (Recency).
\item
  \textbf{Wording:} The exact template string fed to the model during
  the experiment. The \textbf{\{e\}} represents where the specific
  answer word (the entity) is dynamically injected into the text.
\end{itemize}

\newpage

\subsection{Appendix E --- Results table
definitions}\label{appendix-e-results-table-definitions}

\subsubsection{3.1 Positional bias across 10 test
conditions}\label{positional-bias-across-10-test-conditions}

The following table displays the measurement of the model's positional
bias across all ten of the experiment's test conditions:

{\def\LTcaptype{none} 
\begin{longtable}[]{@{}
  >{\centering\arraybackslash}p{(\linewidth - 14\tabcolsep) * \real{0.0800}}
  >{\centering\arraybackslash}p{(\linewidth - 14\tabcolsep) * \real{0.1200}}
  >{\centering\arraybackslash}p{(\linewidth - 14\tabcolsep) * \real{0.1500}}
  >{\centering\arraybackslash}p{(\linewidth - 14\tabcolsep) * \real{0.0800}}
  >{\centering\arraybackslash}p{(\linewidth - 14\tabcolsep) * \real{0.1700}}
  >{\centering\arraybackslash}p{(\linewidth - 14\tabcolsep) * \real{0.1400}}
  >{\centering\arraybackslash}p{(\linewidth - 14\tabcolsep) * \real{0.1300}}
  >{\centering\arraybackslash}p{(\linewidth - 14\tabcolsep) * \real{0.1300}}@{}}
\toprule\noalign{}
\begin{minipage}[b]{\linewidth}\centering
Cell
\end{minipage} & \begin{minipage}[b]{\linewidth}\centering
Header
\end{minipage} & \begin{minipage}[b]{\linewidth}\centering
Body
\end{minipage} & \begin{minipage}[b]{\linewidth}\centering
Cues
\end{minipage} & \begin{minipage}[b]{\linewidth}\centering
Serial Position (half-effect, nats)
\end{minipage} & \begin{minipage}[b]{\linewidth}\centering
95\% CI
\end{minipage} & \begin{minipage}[b]{\linewidth}\centering
Items Showing Primacy
\end{minipage} & \begin{minipage}[b]{\linewidth}\centering
90\% CI clears -0.15
\end{minipage} \\
\midrule\noalign{}
\endhead
\bottomrule\noalign{}
\endlastfoot
T1 & neutral & verbatim & 2 & -0.485 & {[}-0.611, -0.283{]} & 12/13 &
\(\checkmark\) {[}-0.592, -0.318{]} \\
T1c & neutral & verbatim & 0 & \textbf{-0.884} & {[}-1.074, -0.708{]} &
13/13 & \(\checkmark\) {[}-1.044, -0.735{]} \\
T2 & neutral & mixed & 1 & -0.377 & {[}-0.496, -0.227{]} & 12/13 &
\(\checkmark\) {[}-0.478, -0.252{]} \\
T2c & neutral & mixed & 0 & -0.635 & {[}-0.902, -0.452{]} & 13/13 &
\(\checkmark\) {[}-0.854, -0.478{]} \\
T3 & framed & verbatim & 2 & -0.679 & {[}-0.848, -0.520{]} & 13/13 &
\(\checkmark\) {[}-0.822, -0.543{]} \\
T3c & framed & verbatim & 0 & -0.569 & {[}-0.739, -0.429{]} & 13/13 &
\(\checkmark\) {[}-0.713, -0.452{]} \\
T4 & framed & mixed & 1 & \textbf{-0.172} & {[}-0.278, -0.066{]} & 10/13
& \(\times\) {[}-0.262, -0.082{]} \\
T4c & framed & mixed & 0 & -0.482 & {[}-0.603, -0.345{]} & 13/13 &
\(\checkmark\) {[}-0.586, -0.366{]} \\
T5 & framed & loaded & 2 & -0.205 & {[}-0.308, -0.068{]} & 11/13 &
\(\times\) {[}-0.293, -0.091{]} \\
T5b & framed & loaded & 0 & -0.265 & {[}-0.421, -0.137{]} & 10/13 &
\(\checkmark\) \(\dagger\) {[}-0.394, -0.159{]} \\
\end{longtable}
}

\textbf{Definitions:}

\begin{itemize}
\item
  \textbf{Cell:} The specific test condition being evaluated. The
  possible results are the ten conditions created for the study: T1,
  T1c, T2, T2c, T3, T3c, T4, T4c, T5, and T5b.

  Suffixes on the test conditions (such as the `c' in T1c or the `b' in
  T5b) denote the cue-free (0-cue) variants of the base tests. These
  specific identifiers are retained to match the exact condition names
  logged in the open-source repository.
\item
  \textbf{Header:} The wording style used for the labels on the text
  blocks. The possible results are \emph{neutral} (labeled blandly with
  no bias), or one of these framed types: \emph{authority} (official
  guideline) or \emph{recency} (new update). Which one is printed first
  is counterbalanced.
\item
  \textbf{Body:} The wording style used within the actual text blocks.
  The possible results are: \emph{verbatim} (word-for-word identical),
  \emph{mixed} (reworded), or \emph{loaded} (carrying the manipulation
  directly in the sentence).
\item
  \textbf{Cues:} The number of times a specific, exact word pattern
  appears in the text, which researchers tested to see if it triggered a
  copy-and-paste mechanism in the model. The possible results are 0, 1,
  or 2.
\item
  \textbf{Serial Position (half-effect, nats):} This number measures the
  model's positional bias in \emph{nats} (a unit of information). A
  negative number indicates a \emph{primacy effect} (the model favors
  the first block it reads), while a positive number indicates a
  \emph{recency effect} (the model favors the last block).
\item
  \textbf{95\% CI:} The 95\% confidence interval, showing the spread of
  results generated from resampling the test items 9,999 times to ensure
  the effect is reliable. The results are shown as a range.
\item
  \textbf{Items Showing Primacy:} How many of the 13 word pairs (items)
  tested actually exhibited a negative serial number (a primacy effect).
\item
  \textbf{90\% CI clears -0.15:} Answers whether the effect is large
  enough to be considered non-trivial, specifically checking if the 90\%
  confidence interval lies entirely beyond a -0.15 boundary. The results
  are either a checkmark (\(\checkmark\)) along with the interval if it
  clears the boundary, or a cross (\(\times\)) if it fails to clear it,
  with one result receiving a dagger (\(\dagger\)) for passing by a tiny
  margin.
\end{itemize}

\textbf{Group definitions:}

\begin{itemize}
\item
  \textbf{Cell}, \textbf{Header}, \textbf{Body}, and \textbf{Cues}: The
  test conditions. These outline the specific combination wording and
  structural settings used to create a particular test.
\item
  \textbf{Serial Position}: Shows how much the model leaned toward a
  document based solely on where it was placed in the prompt. A negative
  number indicates a primacy effect while a positive number indicates a
  recency effect.
\item
  \textbf{95\% CI}, \textbf{Items Showing Primacy}, and \textbf{90\% CI
  clears -0.15}: Statistical proof to confirm that the measured effect
  was consistent across the tested items and large enough to be
  considered non-trivial.
\end{itemize}

\subsubsection{3.2 Contrasts}\label{contrasts}

The following table displays the specific comparisons (contrasts) used
to test whether removing repeated word patterns (the copy cues) reduces
the model's bias toward the first document:

{\def\LTcaptype{none} 
\begin{longtable}[]{@{}
  >{\raggedright\arraybackslash}p{(\linewidth - 6\tabcolsep) * \real{0.2500}}
  >{\raggedright\arraybackslash}p{(\linewidth - 6\tabcolsep) * \real{0.2500}}
  >{\raggedright\arraybackslash}p{(\linewidth - 6\tabcolsep) * \real{0.2500}}
  >{\raggedright\arraybackslash}p{(\linewidth - 6\tabcolsep) * \real{0.2500}}@{}}
\toprule\noalign{}
\begin{minipage}[b]{\linewidth}\raggedright
Contrast
\end{minipage} & \begin{minipage}[b]{\linewidth}\raggedright
Description
\end{minipage} & \begin{minipage}[b]{\linewidth}\raggedright
Filtered Dataset (13 items, primary)
\end{minipage} & \begin{minipage}[b]{\linewidth}\raggedright
Unfiltered Dataset (14 items)
\end{minipage} \\
\midrule\noalign{}
\endhead
\bottomrule\noalign{}
\endlastfoot
A1 & Cue removal, neutral header (T1c - T1) & -0.400 {[}-0.578,
-0.237{]} * & -0.407 {[}-0.573, -0.251{]} * \\
A2 & Cue removal, authority header (T3c - T3) & +0.110 {[}-0.081,
+0.287{]} & +0.074 {[}-0.112, +0.257{]} \\
A3 & Average of A1 and A2, descriptive only & -0.145 {[}-0.290,
+0.010{]} & -0.166 {[}-0.306, -0.011{]} * \\
A4 & Do A1 and A2 arms differ? & -0.509 {[}-0.687, -0.316{]} * & not
computed \(\ddagger\) \\
\end{longtable}
}

\textbf{Definitions:}

\begin{itemize}
\item
  \textbf{Contrast:} Each row represents a different test measuring what
  happens when the copy cue is deleted. Each test is defined as an arm.

  \begin{itemize}
  \item
    \textbf{A1}: Shows the effect of deleting the cue when the documents
    have neutral, unbiased titles.
  \item
    \textbf{A2:} Tests deleting the cue when the documents have
    official, authoritative titles.
  \item
    \textbf{A3:} Calculates the average between the first two tests (A1,
    A2).

    We did not average the two arms A1 and A2. Statistically, two
    measurements should only be averaged if they capture the same
    underlying effect. Our analysis script automatically tested this by
    calculating the difference between the two conditions. The resulting
    confidence interval excluded zero, proving the two arms differ
    significantly. Consequently, the script issued a warning against
    pooling these heterogeneous results, so we reported A1 and A2
    separately and treated their average (A3) purely as a descriptive
    metric with zero inferential weight.
  \item
    \textbf{A4:} Calculates the difference between the first two tests
    (A1, A2).

    \(\ddagger\) Evaluated on the primary filtered dataset only. Because
    arms A1 and A2 are analyzed individually in Section 3.2, this
    comparison is not computed for the unfiltered sensitivity set.
  \end{itemize}
\item
  \textbf{Filtered Dataset:} The results using the strict, primary
  dataset of 13 word pairs that passed all sanity checks. The result is
  presented in this format:
  \emph{\texttt{contrast\ {[}95\%\ confidence\ interval{]}\ *}}
\item
  \textbf{Unfiltered Dataset:} A sensitivity check where the researchers
  re-ran the calculations with 14 word pairs (including one they had
  previously excluded) to make sure the overall conclusions didn't
  change. The result is presented in this format:
  \emph{\texttt{contrast\ {[}95\%\ confidence\ interval{]}\ *}}
\item
  \textbf{contrast {[}95\% confidence interval{]} *}:

  \begin{itemize}
  \item
    \textbf{contrast:} The difference between two different test
    conditions. A negative number represents a primacy effect, a
    positive number represents a recency effect.
  \item
    \textbf{95\% confidence interval:} Shows the spread of results
    generated by resampling 9,999 times to ensure the measured effect
    was reliable.
  \item
    \textbf{( * ):} The 95\% confidence interval does not include zero.
    When the interval excludes zero, it means the measured effect is
    statistically significant and consistently points in the same
    direction across all the resampled test variations.
  \end{itemize}
\end{itemize}

\subsubsection{3.3 Framing and position
influence}\label{framing-and-position-influence}

The following table contains the test conditions we used to measure
framing and position influence:

{\def\LTcaptype{none} 
\begin{longtable}[]{@{}llll@{}}
\toprule\noalign{}
Test Condition & Header Wording & Body Wording & Cue Count \\
\midrule\noalign{}
\endhead
\bottomrule\noalign{}
\endlastfoot
T5 & framed & loaded & 2 \\
T5b & framed & loaded & 0 \\
\end{longtable}
}

\textbf{Definitions:}

\begin{itemize}
\item
  \textbf{Test Condition:} The specific experimental setup being
  evaluated. The body text acts as the manipulation and is role-bound to
  the header.

  \begin{itemize}
  \item
    \textbf{T5:} Repeated word patterns (copy cues) are present.
  \item
    \textbf{T5b:} All repeated word patterns (copy cues) are completely
    removed.
  \end{itemize}
\item
  \textbf{Header Wording:} The wording style used for the labels on the
  text blocks. In these specific tests, they are \emph{framed} (present
  the document as an official guideline or a fresh update).
\item
  \textbf{Body Wording:} The wording style used within the actual text
  blocks. In these tests, the body is loaded (the authoritative or
  recent tone is embedded directly within the sentences themselves).
\item
  \textbf{Cue Count:} The number of times a specific, exact word pattern
  (e.g.~``is {[}Answer{]}'') appears in the text. We used this to test
  if the model was relying on a simple copy-and-paste mechanism.
\end{itemize}

\subsubsection{3.3 Quantities for hidden
artifacts}\label{quantities-for-hidden-artifacts}

The following table contains the quantities we measured for hidden
artifacts and natural vocabulary biases in T5b:

{\def\LTcaptype{none} 
\begin{longtable}[]{@{}
  >{\raggedright\arraybackslash}p{(\linewidth - 4\tabcolsep) * \real{0.3333}}
  >{\raggedright\arraybackslash}p{(\linewidth - 4\tabcolsep) * \real{0.3333}}
  >{\raggedright\arraybackslash}p{(\linewidth - 4\tabcolsep) * \real{0.3333}}@{}}
\toprule\noalign{}
\begin{minipage}[b]{\linewidth}\raggedright
Quantity
\end{minipage} & \begin{minipage}[b]{\linewidth}\raggedright
Estimate
\end{minipage} & \begin{minipage}[b]{\linewidth}\raggedright
95\% CI
\end{minipage} \\
\midrule\noalign{}
\endhead
\bottomrule\noalign{}
\endlastfoot
Source framing (header + body together) & \textbf{+1.054} & {[}+0.907,
+1.258{]} * \\
Serial position (half-effect) & \textbf{-0.265} & {[}-0.421, -0.137{]}
* \\
Residual check (block-order null control) & +0.011 & \(\checkmark\)
{[}-0.058, +0.087{]} \\
Token (base word preference) & +0.003 & \(\checkmark\) {[}-0.135,
+0.165{]} \\
\end{longtable}
}

\textbf{Definitions:}

\begin{itemize}
\item
  \textbf{Quantity:} The specific variable, force, or sanity check that
  we measured in test condition T5b.

  \begin{itemize}
  \item
    \textbf{Source framing:} How much the authoritative or recent tone
    of the document influenced the model's answer.
  \item
    \textbf{Serial position:} How much the order of the documents (first
    versus last) influenced the model's answer.
  \item
    \textbf{Residual check:} A sanity check. The near-zero Order value
    (+0.011) confirms no hidden positional artifacts leaked into the
    results.
  \item
    \textbf{Token:} A sanity check. The near-zero Token value (+0.003)
    proves that the model's natural vocabulary bias was completely
    canceled out.
  \end{itemize}
\item
  \textbf{Estimate:} The calculated measurement of the effect, expressed
  in nats. It shows how hard the model's decision was pulled in a
  certain direction.
\item
  \textbf{95\% CI:} The 95\% confidence interval. This shows the spread
  of results generated by resampling the test items 9,999 times to
  ensure the measurement is reliable. An asterisk (*) means the effect
  is statistically significant because the range does not include zero.
\end{itemize}

\subsubsection{3.3 Influential factors}\label{influential-factors}

We use the following contrasts to separate the different factors that
influence the model's decisions:

{\def\LTcaptype{none} 
\begin{longtable}[]{@{}
  >{\raggedright\arraybackslash}p{(\linewidth - 6\tabcolsep) * \real{0.2500}}
  >{\raggedright\arraybackslash}p{(\linewidth - 6\tabcolsep) * \real{0.2500}}
  >{\raggedright\arraybackslash}p{(\linewidth - 6\tabcolsep) * \real{0.2500}}
  >{\raggedright\arraybackslash}p{(\linewidth - 6\tabcolsep) * \real{0.2500}}@{}}
\toprule\noalign{}
\begin{minipage}[b]{\linewidth}\raggedright
Contrast
\end{minipage} & \begin{minipage}[b]{\linewidth}\raggedright
Description
\end{minipage} & \begin{minipage}[b]{\linewidth}\raggedright
Filtered Dataset (\(n = 13\))
\end{minipage} & \begin{minipage}[b]{\linewidth}\raggedright
Unfiltered Dataset (\(n = 14\))
\end{minipage} \\
\midrule\noalign{}
\endhead
\bottomrule\noalign{}
\endlastfoot
D1 & Framing vs.~position margin, cue-free (T5b) & +0.774 {[}+0.585,
+0.966{]} * & +0.806 {[}+0.616, +0.991{]} ** \\
D2 & Framing vs.~position margin, with cues (T5) & +0.316 {[}+0.188,
+0.434{]} * & +0.317 {[}+0.199, +0.428{]} * \\
D3 & Effect of removing cues on framing (T5b - T5) & +0.539 {[}+0.330,
+0.843{]} * & +0.583 {[}+0.366, +0.874{]} * \\
\end{longtable}
}

\textbf{Definitions:}

\begin{itemize}
\item
  \textbf{Contrast:} A calculation that measures the exact difference
  between test conditions to separate the various factors influencing
  the model's decisions.

  \begin{itemize}
  \item
    \textbf{D1:} Measures the head-to-head difference between the
    framing effect and the positional effect in the completely cue-free
    condition (T5b).
  \item
    \textbf{D2:} Measures that exact same head-to-head comparison, but
    in the test condition that included copy cues (T5).
  \item
    \textbf{D3:} Calculates the difference in the framing effect between
    the cued test and the cue-free test.
  \end{itemize}
\item
  \textbf{Filtered Dataset:} The results using the strict, primary
  dataset of 13 word pairs that passed all sanity checks. The result is
  presented in this format:
  \emph{\texttt{contrast\ {[}95\%\ confidence\ interval{]}\ *}}
\item
  \textbf{Unfiltered Dataset:} A sensitivity check where the researchers
  re-ran the calculations with 14 word pairs (including one they had
  previously excluded) to make sure the overall conclusions didn't
  change. The result is presented in this format:
  \emph{\texttt{contrast\ {[}95\%\ confidence\ interval{]}\ *}}
\end{itemize}

Note: In all three contrasts, the asterisk (*) confirms that the 95\%
confidence intervals do not include zero, meaning all of these findings
are statistically significant.

\subsubsection{3.4 Authority headers}\label{authority-headers}

To ensure clean results, all three of the comparisons below were
conducted on text completely free of repeated copy cues:

{\def\LTcaptype{none} 
\begin{longtable}[]{@{}
  >{\raggedright\arraybackslash}p{(\linewidth - 6\tabcolsep) * \real{0.2500}}
  >{\raggedright\arraybackslash}p{(\linewidth - 6\tabcolsep) * \real{0.2500}}
  >{\raggedright\arraybackslash}p{(\linewidth - 6\tabcolsep) * \real{0.2500}}
  >{\raggedright\arraybackslash}p{(\linewidth - 6\tabcolsep) * \real{0.2500}}@{}}
\toprule\noalign{}
\begin{minipage}[b]{\linewidth}\raggedright
Contrast
\end{minipage} & \begin{minipage}[b]{\linewidth}\raggedright
Description
\end{minipage} & \begin{minipage}[b]{\linewidth}\raggedright
Filtered Dataset (\(n = 13\))
\end{minipage} & \begin{minipage}[b]{\linewidth}\raggedright
Unfiltered Dataset (\(n = 14\))
\end{minipage} \\
\midrule\noalign{}
\endhead
\bottomrule\noalign{}
\endlastfoot
C1 & Effect of authority header, mixed bodies (T4c - T2c) & +0.153
{[}-0.011, +0.352{]} & +0.123 {[}-0.033, +0.322{]} \\
C2 & Effect of authority header, verbatim bodies (T3c - T1c) & +0.315
{[}+0.172, +0.420{]} * & +0.296 {[}+0.166, +0.403{]} * \\
C3 & The interaction (C1 - C2) & -0.162 {[}-0.303, -0.034{]} * & -0.173
{[}-0.301, -0.048{]} * \\
\end{longtable}
}

\textbf{Definitions:}

\begin{itemize}
\item
  \textbf{Contrast:} A calculation used to measure the effect of an
  authoritative header and test whether that effect changes depending on
  the wording of the body text.

  \begin{itemize}
  \item
    \textbf{C1:} Measures how much an authoritative header changes the
    model's bias when the competing documents are reworded (mixed).
  \item
    \textbf{C2:} Measures how much an authoritative header changes the
    model's bias when the competing documents are word-for-word
    identical (verbatim).
  \item
    \textbf{C3:} Calculates the exact difference between C1 and C2 to
    test whether the body wording significantly alters the header's
    impact.
  \end{itemize}
\item
  \textbf{Filtered Dataset:} The results using the strict, primary
  dataset of 13 word pairs that passed all sanity checks. The result is
  presented in this format:
  \emph{\texttt{contrast\ {[}95\%\ confidence\ interval{]}\ *}}
\end{itemize}

\newpage

\subsection{Glossary}\label{glossary}

\begin{itemize}
\tightlist
\item
  \textbf{Activation patching:} A mechanistic interpretability technique
  used to identify specific neural circuits and attention heads by
  replacing intermediate hidden activations during inference.
\item
  \textbf{AD series:} The final experimental iteration (recorded in
  \texttt{AD\_series\_FINAL.txt}) that resolved design limitations of
  earlier exploratory series (AA through AC) and provides the definitive
  dataset for this paper.
\item
  \textbf{Behavioral suite:} An evaluation framework measuring model
  input--output behavior under controlled prompt variations from the
  outside, without assuming internal representations.
\item
  \textbf{Block order:} The presentation order in which competing
  document blocks are delivered to the model (e.g., whether Block 1 or
  Block 2 appears first).
\item
  \textbf{CI:} Confidence Interval. See \emph{Confidence Interval}.
\item
  \textbf{Confidence Interval:} A statistical range (here, 95\% BCa
  bootstrap with 9,999 resamples) estimating the plausible spread of an
  effect size across test items.
\item
  \textbf{Copy cue:} A specific word pattern suspected of triggering
  mechanical copying. See \emph{Cue}.
\item
  \textbf{Counterbalancing:} Systematically permuting variables
  (document order and entity-role assignments) across configurations so
  baseline biases cancel out by symmetry.
\item
  \textbf{Cue:} A repeated surface word pattern (e.g., \emph{``is
  \{e\}''}, \emph{``is now \{e\}''}) hypothesized to mechanically bias
  model predictions toward an answer.
\item
  \textbf{Entity:} A specific candidate answer noun (e.g., \emph{Scrum},
  \emph{Kanban}, \emph{Agile}, \emph{Waterfall}).
\item
  \textbf{Entity-role assignment:} The mapping determining which
  candidate entity plays which semantic role in a prompt (e.g.,
  assigning \emph{Agile} to the authoritative guideline and
  \emph{Waterfall} to the team update, or vice versa).
\item
  \textbf{Forward pass:} A single evaluation run feeding an input prompt
  through the model to compute next-token output logits.
\item
  \textbf{Four crossings:} The \(2 \times 2\) factorial design crossing
  block orders with entity-role assignments to isolate true behavioral
  effects while canceling out positional and lexical biases (Appendix
  C).
\item
  \textbf{Lexical prior:} See \emph{Vocabulary bias}.
\item
  \textbf{Mechanistic claim:} An assertion attributing a model's
  behavior to specific internal computational structures (e.g.,
  attention heads or circuits), distinct from purely behavioral
  observations of input--output relationships.
\item
  \textbf{Nat:} The standard unit of information based on natural
  logarithms (base \(e\)), where
  \(1\text{ nat} \approx 1.44\text{ bits}\). Used to quantify log-odds
  shifts and log-probability differences in model predictions.
\item
  \textbf{Positional bias:} Systematic preference for an answer based
  purely on its position in a prompt (e.g., primacy or recency),
  independent of content.
\item
  \textbf{Primacy:} Systematic bias favoring information presented
  earlier in a prompt.
\item
  \textbf{Recency:} Systematic bias favoring information presented later
  in a prompt.
\item
  \textbf{Serial position:} The presentation order of documents in
  context; one of the two primary forces measured in this study.
\item
  \textbf{Source framing:} The semantic framing or authority attributed
  to a document (e.g., an official guideline versus a team update); the
  dominant force measured in this study.
\item
  \textbf{Statistical model:} A mathematical model that assigns
  probabilities to token sequences based on patterns learned during
  pretraining.
\item
  \textbf{Surface wording:} The superficial phrasing used to present
  information without altering the underlying factual proposition (e.g.,
  verbatim repetition vs.~paraphrased body text).
\item
  \textbf{Vocabulary bias (lexical prior):} The model's baseline
  preference for one word over another regardless of context (e.g.,
  naturally preferring \emph{Agile} over \emph{Waterfall}).
\end{itemize}

\end{document}